\documentclass[authorversion, nonacm]{acmart}

\AtBeginDocument{%
  \providecommand\BibTeX{{%
    \normalfont B\kern-0.5em{\scshape i\kern-0.25em b}\kern-0.8em\TeX}}}

\copyrightyear{2023}
\acmYear{2023}
\setcopyright{rightsretained}
\acmConference[EAAMO '23]{Equity and Access in Algorithms, Mechanisms, and Optimization}{October 30-November 1, 2023}{Boston, MA, USA}
\acmBooktitle{Equity and Access in Algorithms, Mechanisms, and Optimization (EAAMO '23), October 30-November 1, 2023, Boston, MA, USA}
\acmDOI{10.1145/3617694.3623222}
\acmISBN{979-8-4007-0381-2/23/11}

\usepackage{tikz}
\usetikzlibrary{arrows}
\usetikzlibrary{shapes.geometric}
\usepackage{caption}
\usepackage{subcaption}
\usepackage[algo2e, ruled]{algorithm2e} 
\usepackage{soul}
\usepackage{bbm}
\usepackage{eurosym}

\begin{document}

\title[Counterfactual Situation Testing]{Counterfactual Situation Testing: Uncovering Discrimination under Fairness given the Difference}

\author{Jose M. Alvarez}
\email{jose.alvarez@sns.it}
\orcid{0000-0001-9412-9013}
\affiliation{%
  \institution{Scuola Normale Superiore, University of Pisa}
  \city{Pisa}
  \country{Italy}
}

\author{Salvatore Ruggieri}
\email{salvatore.ruggieri@unipi.it}
\orcid{0000-0002-1917-6087}
\affiliation{%
  \institution{University of Pisa}
  \city{Pisa}
  \country{Italy}
}

\renewcommand{\shortauthors}{Alvarez and Ruggieri}

\begin{abstract} 
    We present counterfactual situation testing (CST), a causal data mining framework for detecting individual discrimination in a dataset of classifier decisions. CST answers the question ``what would have been the model outcome had the individual, or complainant, been of a different protected status?'' in an actionable and meaningful way. It extends the legally-grounded situation testing of \citet{Thanh_KnnSituationTesting2011} by operationalizing the notion of \textit{fairness given the difference} of \citet{Kohler2018CausalEddie} using counterfactual reasoning. In standard situation testing we find for each complainant similar protected and non-protected instances in the dataset; construct respectively a control and test group; and compare the groups such that a difference in decision outcomes implies a case of potential individual discrimination. In CST we avoid this idealized comparison by establishing the test group on the complainant's counterfactual generated via the steps of abduction, action, and prediction. The counterfactual reflects how the protected attribute, when changed, affects the other seemingly neutral attributes of the complainant. Under CST we, thus, test for discrimination by comparing similar individuals within each group but dissimilar individuals across both groups for each complainant. Evaluating it on two classification scenarios, CST uncovers a greater number of cases than ST, even when the classifier
    is counterfactually fair.
\end{abstract}



\received{10 May 2023}
\received[revised]{5 September 2023}
\received[accepted]{20 September 2023}

\maketitle

\section{Introduction}
\label{sec:Introduction}
Automated decision making (ADM) is becoming ubiquitous and its societal discontents clearer \cite{Angwin2016MachineBias, DastinAmazonSexist, Heikkila2022_DutchScnadal}. There is a shared urgency by regulators \cite{USA_AIBill, EU_AIAct} and researchers \cite{Kleinberg2019DiscriminationAgeOfAlgorithms, Ruggieri2023_TrusFairAI} to develop frameworks that can asses these classifiers for potential discrimination based on protected attributes such as gender, race, or religion. Discrimination is often conceived as a causal claim on the effect of the protected attribute over an individual decision outcome \cite{Heckman1998_DetectingDiscrimination,Foster2004}. It is, in particular, a conception based on counterfactual reasoning---what would have been the model outcome if the individual, or \textit{complainant}, were of a different protected status?---where we ``manipulate'' the protected attribute of the individual. \citet{Kohler2018CausalEddie} calls such conceptualization the \textit{counterfactual causal model of discrimination} (CMD). 

Several frameworks for proving ADM discrimination are based on CMD \cite{Kohler2018CausalEddie}. Central to these frameworks is defining ``similar'' instances to the complainant; arranging them based on their protected status into control and test groups; and comparing the decision outcomes of these groups to detect the effect of the protected attribute. Among the available tools \cite{Romei2014MultiSurveyDiscrimination, Karima2020SurveyCausalFairness, DBLP:journals/fdata/CareyW22}, however, there is a need for one that is both \textit{actionable} and \textit{meaningful}. We consider a framework to be actionable if it can rule out random circumstances for the individual discrimination claim as often required by courts (e.g., \cite{Foster2004, EU2018_NonDiscriminationLaw, Nachbar2020algorithmic}), and meaningful if it can account for known links between the protected attribute and all other attributes when manipulating the former as often demanded by social scientists (e.g., \cite{Bonilla1997_RethinkingRace, Sen2016_RaceABundle, Kasirzadeh2021UseMisuse}).
%
 
In this paper we present \textit{counterfactual situation testing} (CST), a causal data mining framework for detecting instances of individual discrimination in the dataset used by a classifier. It combines (structural)\footnote{Not to be confused with counterfactual explanations \cite{Wachter2017Counterfactual} from the XAI literature, which are not based on structural causal models. Other works like \cite{Karimi2021_AlgoRecourse} use ``structural'' to differentiate counterfactuals from counterfactual explanations. In this paper, counterfactuals \textit{are} structural.} counterfactuals \cite{PearlCausality2009, Pearl2016_CausalInference} with situation testing \cite{Thanh_KnnSituationTesting2011, Zhang_CausalSituationTesting_2016}. \textit{Counterfactuals} answer to counterfactual reasoning and are generated via structural causal models. Under the right causal knowledge, counterfactuals reflect at the individual level how changing the protected attribute affects other seemingly neutral attributes of a complainant. \textit{Situation testing} is a data mining method, based on the homonymous legal tool \cite{Bendick2007SituationTesting, Rorive2009_ProvingDiscrimination}. For each complainant, under some search algorithm and distance function for measuring similarity, it finds and compares a control and test group of similar protected and non-protected instances in the dataset, where a difference between the decision outcomes of the groups implies potential discrimination. CST follows the situation testing pipeline with the important exception that it constructs the test group around the complainant's counterfactual instead of the complainant.

\textbf{An illustrative example.} Consider the scenario in Fig.~\ref{fig:KarimiV2} (used later in Section~\ref{sec:Experiments.IllustrativeExample}) where a bank uses a classifier to accept or reject ($\hat{Y}$) individual loan applications based on annual salary ($X_1$) and account balance ($X_2$). Suppose a female applicant ($A=1$) with $x_1= 35000$ and $x_2=7048$ gets rejected and files for discrimination. The bank is using non-sensitive information to calculate $\hat{Y}$, but according to Fig.~\ref{fig:KarimiV2} there is also a known link between $A$ and $\{X_1, X_2\}$ that questions the neutrality of such information. Under situation testing, we would find a number of female (protected) and male (non-protected) instances with similar characteristics to the complainant. The resulting control and test groups to be compared would both have similar $X_1$ and $X_2$ to the complainant. On one hand, comparing multiple instances allows to check whether the complainant's claim is an isolated event or representative of an unfavorable pattern toward female applicants by the model (i.e., actionability). On the other hand, knowing what we know about $A$ and its influence, would it be fair to compare the similar female and male instances? As argued by previous works \cite{Kohler2018CausalEddie, Hu_facct_sex_20}, the answer is no. This \textit{idealized comparison} takes for granted the effect of gender on annual salary and account balance. Under counterfactual situation testing, instead, we would generate the complainant's counterfactual under the causal knowledge provided, creating a ``male'' applicant with a higher $x_1=50796$ and $x_2=13852$, and use it rather than the complainant to find similar male instances. The resulting control and test groups would have different $X_1$ and $X_2$ between them. This disparate comparison embodies \textit{fairness given the difference}, explicitly acknowledging the lack of neutrality when looking at $X_1$ and $X_2$ based on $A$ (i.e., meaningfulness). Here, the control group represents the observed factual world while the test group the hypothetical counterfactual world of the complainant.

In addition, with counterfactual situation testing we propose an actionable extension to \textit{counterfactual fairness} by \citet{Kusner2017CF}, which remains the leading causal fairness framework \cite{Karima2020SurveyCausalFairness}. A classifier is counterfactually fair when the complainant's and its counterfactual's decision outcomes are the same. These are the same two instances used by CST to construct, respectively, the control and test groups, which allows to equip this fairness definition with measures for uncertainty. Hence, CST links counterfactual fairness claims with notions of statistical significance. Further, by looking at the control and test groups rather than the literal comparison of the factual versus counterfactual instances, CST evaluates whether the counterfactual claim is representative of similar instances. Hence, CST detects cases of individual discrimination that are also counterfactually fair, capturing the realistic scenario where a deployed model tends to discriminate when asked to evaluate a borderline instance multiple times.

Based on two case studies using synthetic and real data, we evaluate the CST framework using a \textit{k}-nearest neighbor implementation, k-NN CST, and compare it to its situation testing counterpart, k-NN ST \cite{Thanh_KnnSituationTesting2011}, as well as to counterfactual fairness \cite{Kusner2017CF}. Here, $k$ denotes the number of instances we wish to find for each control and test groups. The experiments show that CST detects a higher number of individual cases of discrimination across the different $k$ sizes. Further, the results also show that individual discrimination can occur even when the model is counterfactually fair. The results hold when dealing with multiple protected attributes as well as different implementation parameters.

Our main contributions with CST are: \textit{(1)} a meaningful and actionable framework for detecting individual discrimination; \textit{(2)} a first operationalization of \textit{fairness given the difference} for discrimination analysis; and \textit{(3)} an actionable extension of counterfactual fairness equipped with confidence intervals. With this in mind, Section~\ref{sec:CausalKnowledge} explores the role of causal knowledge in CST. Section~\ref{sec:CST} presents the CST framework and its k-NN implementation, while Section~\ref{sec:Experiments} showcases CST via two classification scenarios. Section~\ref{sec:Conclusion} concludes the paper.

\subsection{Related Work}
We position CST with current works along the goals of actionability and meaningfulness.
Regarding actionability, when proving discrimination, it is important to insure that the framework accounts for sources of randomness in the decision process. Popular non-algorithmic frameworks---such as natural \cite{Godin2000Orchestra} and field \cite{Bertrand2017_FieldExperimentDiscrimination} experiments, audit \cite{Fix&Struyk1993_ClearConvincingEvidence} and correspondence \cite{Bertrand2004_EmilyAndGreg, Rooth2021} studies---address this issue by using multiple observations to build inferential statistics.
Similar statistics are sometimes asked in court for proving discrimination (e.g., \cite[Section 6.3]{EU2018_NonDiscriminationLaw}).
Few algorithmic frameworks, instead, address this issue due to model complexity preventing formal inference \cite{Athey2019MachineLearningForEconomists}.
An exception are data mining frameworks for discrimination discovery \cite{Kdd_Pedreschi2008_DataMiningDiscrimination, Ruggieri2010_DMforDD} that operationalize the non-algorithmic notions, including situation testing \cite{Thanh_KnnSituationTesting2011, Zhang_CausalSituationTesting_2016}.
These frameworks (e.g., \cite{TR-DBLP:conf/sigsoft/GalhotraBM17, TR-DBLP:journals/corr/abs-1809-03260, DBLP:journals/jiis/QureshiKKRP20}) keep the focus on comparing multiple control-test instances for making individual claims, providing evidence similar to that produced by the quantitative tools used in court \cite{Kohler2018CausalEddie}.
%
%
To the best of our knowledge, it remains unclear if the same can be said about existing causal fair machine learning methods \cite{Karima2020SurveyCausalFairness} as these have yet to be used beyond academic circles. 

Regarding meaningfulness, situation testing and the other methods have been criticized for their handling of the counterfactual question behind the causal model of discrimination \cite{Kohler2018CausalEddie, Hu_facct_sex_20, Kasirzadeh2021UseMisuse}. In particular, these actionable methods take for granted the influence of the protected attribute on all other attributes. 
This can be seen, e.g., in how situation testing constructs the test group, which is equivalent to changing the protected attribute while keeping everything else equal. Such approach goes against how most social scientists interpret the protected attribute and its role as a social construct when proving discrimination \cite{Bonilla1997_RethinkingRace, rose_constructivist_2022, Sen2016_RaceABundle, Hanna2020_CriticalRace}. It is in that regard where structural causal models \cite{PearlCausality2009} and their ability for conceiving counterfactuals (e.g., \cite{Chiappa2019_PathCF, Yang2021_CausalIntersectionality}), including counterfactual fairness \cite{Kusner2017CF}, have an advantage. What the criticisms on counterfactuals \cite{Hu_facct_sex_20, Kasirzadeh2021UseMisuse} overlook here is that generating counterfactuals, as long as the causal knowledge is properly specified, accounts modeling-wise for the effects of changing the protected attribute on all other observed attributes. A framework like counterfactual fairness, relative to situation testing and these other methods, is more meaningful in its handling of protected attributes.
%
The novelty in CST is bridging these two lines of work, borrowing the actionability aspects from situation testing and the meaningful aspects from counterfactual fairness. 
%

%
%

%
%

\section{Causal Knowledge for Discrimination}
\label{sec:CausalKnowledge}
Counterfactual situation testing requires access to the dataset of decision records of interest, $\mathcal{D}$, and the algorithmic decision-maker that produced it, $b()$. Let $\mathcal{D}$ contain the set of relevant attributes $X$, the set of protected attributes $A$, and the decision outcome $\hat{Y}=b(X)$. We describe $\mathcal{D}$ as a collection of $n$ tuples, each $(x_i, a_i, \widehat{y}_i)$ representing the $i^{th}$ individual profile, with $i \in [1, n]$. $\hat{Y}$ is binary with $\hat{Y} = 1$ denoting the positive outcome (e.g., loan granted). For illustrative purposes, we assume a single binary $A$ with $A=1$ denoting the protected status (e.g., female), though we relax this assumption in the experiments of Section~\ref{sec:Experiments.Real}. We also require causal knowledge in the form of a structural causal model that describes the data generating model behind $\mathcal{D}$. We view this requirement as an input space for experts as these models are a convenient way for organizing assumptions on the source of the discrimination, facilitating stakeholder participation and supporting collaborative reasoning about contested concepts \cite{Mulligan2022_AFCP}.   

\subsection{Structural Causal Models and Counterfactuals}
\label{sec:CausalKnowledge.SCM}

A \textit{structural causal model} (SCM) \cite{PearlCausality2009} $\mathcal{M}=\{ \mathcal{S}, \mathcal{P}_{\mathbf{U}} \}$ describes how the set of $p$ variables $W = X \cup A$ is determined based on corresponding sets of structural equations $\mathcal{S}$ and latent variables $U$ with prior distribution $\mathcal{P}_{\mathbf{U}}$. Each $W_j \in W$ is assigned a value through a deterministic function $f_j \in \mathcal{S}$ of its causal parents $W_{pa(j)} \subseteq W \setminus \{ W_j \}$ and noise variable $U_j$ with distribution $P(U_j) \in \mathcal{P}_{\mathbf{U}}$. Formally, for $W_j \in W$ we have that $W_j \leftarrow f_j(W_{pa(j)}, U_j)$, indicating the flow of information in terms of child-parent or cause-effect pairs. We consider the associated \textit{causal graph} $\mathcal{G} = (\mathcal{V}, \mathcal{E})$, where a node $V_j \in \mathcal{V}$ represents a $W_j$ variable and a directed edge $E_{(j, j')} \in \mathcal{E}$ a causal relation.

We make two assumptions on $\mathcal{M}$ common within the causal fairness literature \citep{Karima2020SurveyCausalFairness}. First, we assume \textit{causal sufficiency}, meaning there are no hidden common causes in $\mathcal{M}$, or confounders. Second, we assume $\mathcal{G}$ to be \textit{acyclical}, which turns $\mathcal{G}$ into a directed acyclical graph (DAG), allowing for no feedback loops.
We write $\mathcal{M}$ under these assumptions as:
\begin{equation}
\label{eq:SCM}
    \mathcal{M} = (\mathcal{S}, \mathcal{P}_{U}), \;\;\;
    \mathcal{S} = \{W_j \leftarrow f_j(W_{pa(j)}, U_j) \}_{j=1}^p, \;\;\;
    \mathcal{P}_{\mathbf{U}} = P(U_1) \times \dots \times P(U_p)
\end{equation}
where these assumptions are necessary for generating counterfactuals. The causal sufficiency assumption is particularly deceitful as it is difficult to both test and account for a hidden confounder \cite{DBLP:journals/corr/abs-1902-10286, mccandless2007bayesian, DBLP:conf/nips/LouizosSMSZW17}. 
The risk of a hidden confounder is a general problem to modeling fairness. Here, the dataset $\mathcal{D}$ delimits our context. We expect it to contain all relevant information used by the decision-maker $b()$.

Input from several stakeholders is needed to derive \eqref{eq:SCM}. We see it as a necessary collaborative effort: \textit{before we implement CST, we first need to agree over a worldview for the discrimination context}. Based on $\mathcal{D}$, a domain-expert motivates a causal graph $\mathcal{G}$. A modelling-expert then translates this graphical information into a SCM $\mathcal{M}$, making model specification decisions on $\mathcal{S}$. We do not cover this process here, but this is how we envision the initial implementation stage of counterfactual situation testing.

For a given SCM $\mathcal{M}$ we want to run \textit{counterfactual queries} to build the test group for a complainant. Counterfactual queries answer to \textit{what would have been if} questions. In CST, we wish to ask such questions around the protected attribute $A$, by setting $A$ to the non-protected status $\alpha$ using the \textit{do-operator} $do(A := \alpha)$ \cite{PearlCausality2009} to capture the individual-level effects $A$ has on $X$ according to \eqref{eq:SCM}. Let $X^{CF}$ denote the set of counterfactual variables obtained via the three step procedure by \citet{Pearl2016_CausalInference}. \textit{Abduction}: for each prior distribution $P(U_i)$ that describes $U_i$, we compute its posterior distribution given the evidence, or $P(U_i \;| \; X, A)$. \textit{Action}: we intervene $A$ by changing its structural equation to $A := \alpha$, which gives way to a new SCM $\mathcal{M}'$. \textit{Prediction}: we generate the \textit{counterfactual distribution} $P(X^{CF}_{A \leftarrow \alpha}(U) \; | \; X, A)$ by propagating the abducted $P(U_i \;| \; X, A)$ through the revised structural equations in $\mathcal{M}'$. 

Note that generating counterfactuals and, thus, CST, unlike, e.g., counterfactual explanations \cite{Wachter2017Counterfactual} and discrimination frameworks like the FlipTest \cite{BlackYF20_FlipTest}, does not require a change in the individual decision outcome. It is possible for $\widehat{Y} = \widehat{Y}^{CF}$ after manipulating $A$. 
We include a working example on generating counterfactuals in the Appendix.

\subsection{Conceiving Discrimination}
\label{sec:CausalKnowledge.IndDisc}

The legal setting of interest is indirect discrimination under EU law. It occurs when an apparently neutral practice disadvantages individuals that belong to a protected group. Following \cite{Hacker2018TeachingFairness}, we focus on indirect discrimination for three reasons. First, unlike disparate impact under US law \cite{Barocas2016_BigDataImpact}, the decision-maker can still be liable for it despite lack of premeditation and, thus, all practices need to consider potential indirect discrimination implications. Second, many ADM models are not allowed to use the protected attribute as input, making it difficult for regulators to use the direct discrimination setting.
Third, we conceive discrimination as a product of a biased society where $b()$ continues to perpetuate the bias reflected in $\mathcal{D}$ because it cannot escape making a decision based on $X$ to derive $\hat{Y}$. 

We view the indirect setting as the one that best describes how biased information can still be an issue for an ADM that never uses the protected attribute. Previous causal works \cite{Kilbertus2017AvoidDiscCau, Chiappa2019_PathCF, Plecko2022_CFA} have focused more on whether the paths between $A$ and $\hat{Y}$ are direct or indirect. Here, the causal setting is much simpler. We know that $b()$ only uses $X$, and are more interested in how information from $A$ is carried by $X$ and how can we account for these links using causal knowledge. That said, this does not mean that CST cannot be implemented in other discrimination settings. We simply acknowledge that it was developed with the EU legal framework in mind. 
Proxy discrimination \cite{Tschantz2022_ProxyDisc}, e.g., is one setting that overlaps with the one we have considered.

Finally, we note that an open legal concern for CST is detecting algorithmic discrimination for various protected attributes, or $|A| > 1$. Two kinds of discrimination, \textit{multiple} and \textit{intersectional}, can occur. Consider, e.g., a black female as the complainant. On what protected attribute is she being potentially discriminated on? In multiple discrimination, we would need to detect separately whether the complainant was discriminated as a black and female individual. In intersectional discrimination, we would instead need to detect simultaneously if the complainant was discriminated as a black-female individual. Only multiple discrimination is currently recognized by EU law, which is an issue as an individual can be free from multiple discrimination but fall victim of intersectional discrimination \cite{Xenidis2020_TunningEULaw}.  CST can account operationally for both scenarios.

\subsection{Fairness given the Difference: the Kohler-Hausmann Critique}
\label{sec:CausalKnowledge.KHC}

Here, we make the case---very briefly---that the causal knowledge required for CST makes it a meaningful framework with respect to situation testing \cite{Thanh_KnnSituationTesting2011, Zhang_CausalSituationTesting_2016} and other tools \cite{Romei2014MultiSurveyDiscrimination} for detecting discrimination. The reference work is \citet{Kohler2018CausalEddie}. We refer to the phrase \textit{fairness given the difference},\footnote{A phrase by Kohler-Hausmann during a panel discussion at NeurIPS 2021 workshop on `Algorithmic Fairness through the Lens of Causal Reasoning.} which best captures her overall critique toward the causal model of discrimination, as the \textit{Kohler-Hausmann Critique} (KHC). CST aims to be meaningful by operationalizing the KHC. It builds the test group on the complainant's counterfactual, letting $X^{CF}$ reflect the effects of changing $A$ instead of assuming $X = X^{CF}$. This is because we view the test group as a representation of the hypothetical counterfactual world of the complainant.

As argued by \cite{Kohler2018CausalEddie} and others before \cite{Bonilla1997_RethinkingRace, Sen2016_RaceABundle}, it is difficult to deny that most protected attributes, if not all of them, are \textit{social constructs}. That is, these attributes were used to classify \textit{and} divide groups of people in a systematic way that conditioned the material opportunities of multiple generations \cite{Mallon2007SocialConstruction, rose_constructivist_2022}. Thus, \textit{recognizing $A$ as a social constructs means recognizing that its effects can be reflected in seemingly neutral variables in $X$}. It is recognizing that $A$, the attribute, cannot capture alone the meaning of belonging to $A$ and that we might, as a minimum, have to link it with other attributes to better capture this, such as $A \rightarrow X$ where $A$ and $X$ change in unison. These attributes \textit{summarize the historical processes that fairness researchers are trying to address} today and should not be treated lightly.\footnote{A clear example of this would be the use of race by US policy makers during the early post-WWII era. See, e.g., the historical evidence provided by \citet{Rothstein2017Color} (for housing), \citet{Schneider2008Smack} (for narcotics), and \citet{Adler2019MurderNewOrleansJimCrow} (for policing).}

The notion of \textit{fairness given the difference} centers on how $A$ is treated in the counterfactual causal model of discrimination (CM). The critique goes beyond the standard manipulation concern \cite{Angrist2008MostlyHarmless} in which $A$ is an inmutable attribute. Instead, granted that we \textit{can} or, more precisely, \textit{have to} manipulate $A$ for running a discrimination analysis, the critique goes against how most discrimination frameworks operationalize such manipulation. The KCH emphasizes that \textit{when $A$ changes, $X$ should change as well}. 

Based on KHC, we consider two types of manipulations that summarize existing frameworks. The \textit{ceteris paribus} (CP), or all else equal, manipulation in which $A$ changes but $X$ remains the same. Examples of it include situation testing \cite{Thanh_KnnSituationTesting2011, Zhang_CausalSituationTesting_2016} but also, e.g., the famous correspondence study by \citet{Bertrand2004_EmilyAndGreg}. The \textit{mutatis mutandis} (MM), or changing what needs to be changed, manipulation in which $X$ changes when we manipulate $A$ based on some additional knowledge, like a structural causal model, that explicitly links $A$ to $X$. Counterfactual fairness \cite{Kusner2017CF}, e.g, uses this manipulation. The MM is clearly preferred over the CP manipulation when we view $A$ as a social construct.  
   
%
%

\section{Counterfactual Situation Testing}
\label{sec:CST}
The objective of CST is to \textit{construct} and \textit{compare} a control and test group for each $c$ protected individual, or \textit{complainant}, in $\mathcal{D}$ in a meaningful and actionable way. Let $(x_c, a_c, \widehat{y}_c) \in \mathcal{D}$ denote the \textit{tuple of interest} on which the individual discrimination claim focuses on, where $c \in [1,n]$. We assume access to the ADM $b()$, the dataset $\mathcal{D}$, and a structural causal model $\mathcal{M}$ describing the discrimination context.

There are three key inputs to consider: the \textit{number of instances per group}, $k$; the \textit{similarity distance function of choice}, $d$; and the \textit{strength of the evidence for rejecting the discrimination claim}, $\alpha$. A fourth key input that we fix in this paper is the \textit{search algorithm of choice}, $\phi$, which we set as the \textit{k-nearest neighbors algorithm} (k-NN) \cite{Hastie2009_ElementsSL}. We do so as the k-NN is intuitive, easy to implement, and commonly used by existing situation testing frameworks. We discuss the CST implementation used in Section~\ref{sec:CST.AnImplementation}, including the choice of $d$. 

\subsection{Building Control and Test Groups}
\label{sec:CST_ControlTest}

For complainant $c$, the control and test groups are built on the \textit{search spaces} and \textit{search centers} for each group. The search spaces are derived and, thus, delimited by $\mathcal{D}$: we are looking for individuals that have gone through the same decision process as the complainant. The search centers, however, are derived separately: the one for the control group comes from $\mathcal{D}$, while the one for the test group comes from the corresponding \textit{counterfactual dataset} $\mathcal{D}^{CF}$. The test search center represents the \textit{what would have been if} of the complainant under a \textit{mutatis mutandis} (MM) manipulation of the protected attribute $A$ that motivates the discrimination claim.

\begin{definition}[Search Spaces]
\label{def:SearchSpaces}
    Under a binary $A$, where $A=1$ denotes the protected status, we partition $\mathcal{D}$ into the \textit{control search space} $\mathcal{D}_c=\{(x_i, a_i, \widehat{y}_i) \in \mathcal{D}: a_i=1\}$ and the \textit{test search space} $\mathcal{D}_t=\{(x_i, a_i, \widehat{y}_i) \in \mathcal{D}: a_i=0\}$.
\end{definition}
\begin{definition}[Counterfactual Dataset]
\label{def:CounterDataset}
    The counterfactual dataset $\mathcal{D}^{CF}$ represents the counterfactual mapping of each instance in the dataset $\mathcal{D}$, with known decision maker $b()$ and SCM $\mathcal{M}$, via the abduction, action, and prediction steps \cite{Pearl2016_CausalInference} when setting a binary $A$ to the non-protected value, or \textit{do}($A:=0$).
\end{definition}

To obtain $\mathcal{D}^{CF}$, we consider an SCM $\mathcal{M}$, as in \eqref{eq:SCM}, where $A$ has no causal parents, or is a root node, $A$ affects only the elements of $X$ considered by the expert(s), and $\hat{Y}=b(X)$. Therefore, when generating the counterfactuals on $A$ (Section~\ref{sec:CausalKnowledge.SCM}), under the indirect discrimination setting (Section~\ref{sec:CausalKnowledge.IndDisc}), the resulting $X^{CF}$ in $\mathcal{D}^{CF}$ should reflect an MM manipulation (Section~\ref{sec:CausalKnowledge.KHC}). Under this structural representation, if $A$ changes then $X$ changes too. See, e.g., Fig.~\ref{fig:KarimiV2} and Fig.~\ref{fig:LawSchool}. The counterfactual dataset represents the world that the complainants would have experienced under $A=0$ \textit{given our worldview}. All three definitions extend to $|A| > 1$.

%
\begin{definition}[Search Centers]
\label{def:SearchCenters}
    For a complainant $c$, we use $x_c$ from the tuple of interest $(x_c, a_c, \hat{y}_c) \in \mathcal{D}$ as the control search center for exploring $\mathcal{D}_c \subset \mathcal{D}$, and use $x_c^{CF}$ from the tuple of interest's generated counterfactual $(x_c^{CF}, a_c^{CF}, \hat{y}_c^{CF}) \in \mathcal{D}^{CF}$ as the test search center for exploring $\mathcal{D}_t \subset \mathcal{D}$. 
\end{definition}

Given the factual $\mathcal{D}$ and counterfactual $\mathcal{D}^{CF}$ datasets, we construct the control and test groups for $c$ using the k-NN algorithm under some distance function $d(x, x')$ to measure similarity between two tuples $x$ and $x'$. We want each group or neighborhood to have a size $k$. For the \textit{control group} (\textit{k-ctr}) we use the (factual) tuple of interest $(x_c, a_c, \hat{y}_c) \in \mathcal{D}$ as search center to explore the protected search space $\mathcal{D}_c$:
\begin{equation}
\label{eq:kctr}
    \text{\textit{k-ctr}} = \{ (x_i, a_i, \widehat{y}_i) \in \mathcal{D}_c: rank_{d}( x_c, x_i) \leq k \}
\end{equation}
where $rank_{d}(x_c, x_i)$ is the rank position of $x_i$ among tuples in $\mathcal{D}_c$ with respect to the ascending distance $d$ from $x_c^{CF}$. For the \textit{test group} (\textit{k-tst}) we use the counterfactual tuple of interest $(x^{CF}_c, a^{CF}_c, \widehat{y}^{CF}_c) \in \mathcal{D}^{CF}$ as search center to explore the non-protected search space $\mathcal{D}_t$:
\begin{equation}
\label{eq:ktst}
    \text{\textit{k-tst}} = \{ (x_i, a_i, \widehat{y}_i) \in \mathcal{D}_t: rank_{d}( x^{CF}_c, x_i) \leq k \}
\end{equation}
where $rank_{d}(x_c^{CF}, x_i)$ is the rank position of $x_i$ among tuples in $\mathcal{D}_t$ with respect to the ascending distance $d$ from $x_c^{CF}$. We use the same distance function $d$ for each group. Neither $A$ nor $\hat{Y}$ are used for constructing the groups. Both \eqref{eq:kctr} and \eqref{eq:ktst} can be expanded by including additional constraints, such as a maximum allowed distance. 

The choice of search centers (Def.~\ref{def:SearchCenters}) is what operationalizes \textit{fairness given the difference} for counterfactual situation testing, making it a \textit{meaningful framework} for testing individual discrimination. To build \textit{k-ctr} and \textit{k-tst} using, respectively, $x_c$ and $x_c^{CF}$ is a statement on how we perceive \textit{within group ordering} as imposed by the protected attribute $A$. This is because the search centers must reflect the $A$-specific ordering of the search spaces that each center targets.
Let us consider our illustrative example from Section~\ref{sec:Introduction}.
If being a female ($A=1$) in this society imposes certain systematic limitations that hinder $x_c$, then comparing $c$ to other female instances in the protected search space preserves the group ordering prescribed by $X|A=1$ as all instances involved experience $A$ in the same way. Therefore, given our worldview, the generated counterfactual male instance for $c$ should then reflect the group ordering prescribed by $X|A=0$. We expect $x_c \neq x_c^{CF}$ given what we know about the effects of $A$ on $X$. Using $x_c^{CF}$ as the test search center would allow us to compare $c$ to other male tuples in the non-protected search space without having to reduce $A$ to a phenotype.

One way to look at the previous statement is by considering the notion of effort. If being female requires a higher individual effort to achieve the same $x_c$, then it is fair to compare $c$ to other female instances. However, it is unfair to compare $c$ to other male instances without adjusting for the extra effort not incurred by the male instances for being males. The counterfactual $x_c^{CF}$ should reflect said adjustment. 
See \cite{Chzhen2020WassersteinBarycenters, Chzhen2022MiniMax} on a similar, more formal critique on individual fairness \cite{DworkHPRZ12} notions. 

\subsection{Detecting Discrimination}
\label{sec:CST_Disc}

For a complainant $c$, we compare the control and test groups by looking at the \textit{difference in proportion of negative decision outcomes}, or $\Delta p = p_c - p_t$, such that:
\begin{align}
    \label{eq:p1_and_p2}
    p_c & = \frac{|\{ (x_i, a_i, \widehat{y}_i) \in \text{\textit{k-ctr}}: \hat{y}_i = 0 \}|}{k}
    &
    p_t & = \frac{|\{ (x_i, a_i, \widehat{y}_i) \in \text{\textit{k-tst}}: \hat{y}_i = 0 \}|}{k} 
\end{align}
where $p_c$ and $p_t$ represents the count of tuples with a negative decision outcome ($\hat{Y}=0$), respectively, in the control and test group. Note that only $\hat{Y}$ is used for deriving the proportions. We compute $\Delta p$ for all protected tuples in $\mathcal{D}$ regardless of the their decision outcome $\hat{Y}$.

CST has the option to include or exclude the search centers when calculating \eqref{eq:p1_and_p2}. If we exclude them, then $p_c$ and $p_t$ remain as is; if we include them, then $\hat{y}_c$ and $\hat{y}_c^{CF}$ are counted in $p_c$ and $p_t$, leading to a denominator in both of $k + 1$. We add this option to be able to compare CST against standard situation testing \cite{Thanh_KnnSituationTesting2011, Zhang_CausalSituationTesting_2016}, which excludes the search centers, and counterfactual fairness \cite{Kusner2017CF}, which only uses the search centers.

Since $\Delta p$ is a proportion comparison, it is known to be \textit{asymptotically normally distributed}, which allows to build \textit{Wald confidence intervals} (CI) around it \cite{Thanh_KnnSituationTesting2011}. 
Let $z_{\alpha/2}$ 
be the $1-\alpha/2$ quantile of the standard normal distribution $\mathcal{N}$ for a \textit{significance level} of $\alpha$ (or, conversely, a \textit{confidence level} $(1 - \alpha) \cdot 100$\%). We write the two-sided CI for $\Delta p$ of $c$ as:
\begin{align}
\label{eq:CIs}
    [\Delta p - w_{\alpha}, \Delta p +  w_{\alpha}],
    & \; \; \; \text{with} \; \; \;
    w_{\alpha} =  z_{\alpha/2} \sqrt{\frac{p_c(1 - p_c) - p_t(1 - p_t)}{k}}
\end{align}
The confidence interval \eqref{eq:CIs} responds to the hypothesis that there is individual discrimination, providing a measure of certainty on $\Delta p$ through a range of possible values. For a given claim, if the CI contains the \textit{minimum accepted deviation} $\tau$, we cannot reject the hypothesis of no discrimination with $(1 - \alpha)\cdot 100$\% confidence. In other words, the \textit{null hypothesis} $H_0: \pi = \tau$ cannot be rejected in favor of the \textit{alternative hypothesis} $H_1: \pi > \tau$, where $\pi$ is the true difference in proportion of negative decision outcomes. $\tau$, with a default choice of $\tau = 0$, represents the minimum amount of difference between $p_c$ and $p_t$ that we need to observe to claim individual discrimination. The overall choice of $\alpha$ and $\tau$ will depend on the context of the discrimination claim. It can be motivated, for instance, by legal requirements (set, e.g., by the court \cite{Thanh_KnnSituationTesting2011}), or technical requirements (set, e.g., via power analysis \cite{Cohen2013StatisticalPower}), or both.

\begin{definition}[Individual Discrimination]
\label{def:IndDisc}
    There is potential\footnote{Or \textit{prima facie}. In practice, discrimination needs to be argued against/for. CST alone, as with any other discrimination analysis tool, cannot claim to prove discrimination. It can, however, provide evidence against/for a discrimination case \cite{Romei2014MultiSurveyDiscrimination}.} individual discrimination toward the complainant $c$ if $\Delta p = p_c - p_t > \tau$, meaning the negative decision outcomes rate for the control group is greater than for the test group by some minimum deviation $\tau \in \mathbb{R}^{+}$.
\end{definition}

We do not view Def.~\ref{def:IndDisc} as a matter of individual versus group fairness. When we test whether $b()$ discriminates against $c$, we inevitably pass judgement onto the classifier $b()$ in fear that this behaviour has happened before. In $\mathcal{D}$ we have more than one potential discrimination claim to consider under CST, allowing to draw individual-level conclusions while motivating group-level ones. If $b()$ discriminated against $c$, it also discriminated against what $c$ represents in terms of membership to $A$.

\begin{definition}[Confidence on the Individual Discrimination Claim]
\label{def:CIs}
    The Wald CI \eqref{eq:CIs} gives a measure of certainty on $\Delta p$, which is (asymptotically) normally distributed. For a significance level $\alpha$, we are $(1-\alpha)$\% confident on $\Delta p$. The claim is said to be statistically valid if the Wald CI excludes $\tau$. This is a statistical inference extension of Def.~\ref{def:IndDisc}.
\end{definition}

The many-to-many comparison behind $\Delta p$ is what makes counterfactual situation testing an \textit{actionable framework} for testing individual discrimination. Here, the notion of repetition and its relation to representativeness and certainty concerns is important. For proving individual discrimination a single comparison is not enough \cite[Sec. 6.3]{EU2018_NonDiscriminationLaw}. This is because we want to ensure, one, that the individual claim is representative of the population, and two, be certain about the individual claim. Implicit to both concerns is finding a pattern of unfavorable decisions against the protected group to which the individual complainant belongs to, i.e., discrimination. 

Ideally, we would repeat the decision process multiple times for the discriminatory pattern to become apparent. This is not possible in practice. Back to our illustrative example from Section~\ref{sec:Introduction}, we cannot ask the female complainant to apply multiple times to the same bank. We instead can look at other similar instances under the same process. This is what $p_c$ and $p_t$ \eqref{eq:p1_and_p2} and Def.~\ref{def:IndDisc} represent. Similarly, if the bank's $b()$ is shown to discriminate against the female complainant, what rules out that it has not done it before or that this one time was an exception? Again, we cannot repeat the decision process until we are certain of the individual discrimination claim. We instead can assume a theoretical distribution of comparisons with $\pi$ to account for potential randomness in what we detect from the single point estimate that is $\Delta p$. This is what the CI \eqref{eq:CIs} and Def.~\ref{def:CIs} represent.   

\subsection{Connection to Counterfactual Fairness}
\label{sec:CST.OnCF}

There is a clear link between CST and \textit{counterfactual fairness} \cite{Kusner2017CF}. A decision maker is counterfactually fair if it outputs the same outcome for the factual tuple as for its counterfactual tuple, where the latter is generated based on the abduction, action, and prediction steps and the intervention on the protected attribute.\footnote{Formally, $P(\hat{Y}_{A \leftarrow a}(U)=y \, | \, X, A) = P(\hat{Y}_{A \leftarrow a'}(U)=y \, | \, X, A)$, where the left side is the factual $A=a$ and the right side the counterfactual $A=a'$.} The factual $(x_c, a_c, \hat{y}_c)$ and counterfactual $(x_c^{CF}, a_c^{CF}, \hat{y}_c^{CF})$ tuples for $c$ used in CST are also the ones used for counterfactual fairness. We view CST, when including the search centers, as an actionable extension of counterfactual fairness.

\begin{proposition}[On Actionable Counterfactual Fairness]
\label{prop:ActioanbleCF} 
    Counterfactual fairness does not imply nor it is implied by Individual Discrimination (Def.~\ref{def:IndDisc}).
\end{proposition}

We present a sketch of proof to Prop.~\ref{prop:ActioanbleCF} the in Appendix. Intuitively, it is possible to handle \textit{borderline cases} where the tuple of interest and its counterfactual both get rejected by $b()$, though the latter is close to the decision boundary. The model $b()$ would be considered counterfactually fair, but would that disprove the individual discrimination claim? CST, by constructing the control and test groups around this single comparison, accounts for this actionability concern. CST further equips counterfactual fairness with confidence intervals. Previous works have addressed uncertainty in counterfactual fairness \cite{RussellKLS2017_WorldsCollide, DBLP:conf/uai/KilbertusBKWS19}, but with a focus on the structure of the SCM $\mathcal{M}$. We instead address certainty on the literal comparison that motivates the counterfactual fairness definition.

\subsection{An Implementation: k-NN CST}
\label{sec:CST.AnImplementation}

Finally, we propose an implementation to our counterfactual situation testing framework. We already defined the search algorithm $\phi$ as the k-NN algorithm. We define as the similarity measure $d$ the same distance function between two tuples, $d(x, x')$, used in the k-NN situation testing implementation (k-NN ST) \cite{Thanh_KnnSituationTesting2011}. We do so because we want to compare our implementation, k-NN CST, against its standard counterpart, k-NN ST. We summarize the current algorithmic CST implementation in the Appendix. Let us define the \textit{distance between two tuples} as:
\begin{equation}
\label{eq:Distance}
    d(x, x') = \frac{\sum_{i=1}^{|X|} d_i(x_i - x'_{i})}{|X|}
\end{equation}
where \eqref{eq:Distance} averages the sum of the per-attribute distances across $X$. A lower \eqref{eq:Distance} implies a higher similarity between the tuples $x$ and $x'$. Here, $d_i$ equals the overlap measurement ($ol$) if the attribute $X_i$ is categorical; otherwise, it equals the normalized Manhattan distance ($md$) if the attribute $X_i$ is continuous, ordinal, or interval. We define each $md(x_i, x_{i'}) = {| x_i - x_{i'} |}/{(\max(X) - \min(X))}$, and $ol(x_i, x_{i'}) = 1$ if $x_i = x_{i'}$ and $0$ otherwise.
CST can handle non-normalized attributes but, unless specified, we normalize them to insure comparable per-attribute distances.
The choice of \eqref{eq:Distance} is not restrictive.
In subsequent works we hope to explore other distance options like , e.g., heterogeneous distance functions \cite{WilsonM97_HeteroDistanceFunctions}, as well as probability-based options, e.g., propensity score weighting \cite{DBLP:journals/jiis/QureshiKKRP20}. 

CST is, above all, a framework for detecting discrimination. The choice of $d$ as well as $\phi$ are specific to the implementation of CST. What is important is that the test group is established around the complainant's counterfactual while the control group, like in other discrimination frameworks, is established around the complainant. 

%
%

\section{Experiments}
\label{sec:Experiments}
We now showcase the counterfactual situation testing (CST) framework via its k-NN implementation using synthetic (Section~\ref{sec:Experiments.IllustrativeExample}) and real (Section~\ref{sec:Experiments.Real}) datasets. We contrast it to its situation testing counterpart (k-NN ST) \cite{Thanh_KnnSituationTesting2011}, and to counterfactual fairness (CF) \cite{Kusner2017CF}. Here, for the structural equations $\mathcal{S}$ we assume additive noise. This is a convenient but not necessary assumption that simplifies the abduction step when generating the counterfactuals.\footnote{Formally, we assume $\mathcal{S} = \{W_j \leftarrow f_j(W_{pa(j)}) + U_j \}_{j=1}^p$ for $W = X \cup A$.} We use a significance level of $\alpha=5\%$, a minimum deviation of $\tau=0.0$, and a set of $k$ group sizes in $\{15, 30, 50, 100\}$ for CST runs that include and exclude the search centers. Also for comparison, we define individual discrimination as $\Delta p > \tau$ (Def.~\ref{def:IndDisc}) for a single protected attribute. We still, though, demonstrate the use of confidence intervals (Def.~\ref{def:CIs}) and how it would affect the final results. 
%
%
Finally, 
we assume $\mathcal{M}$ and $\mathcal{G}$ in both ADM scenarios.\footnote{The code and data are available at 
\href{https://github.com/cc-jalvarez/counterfactual-situation-testing}{https://github.com/cc-jalvarez/counterfactual-situation-testing}.}
 
\subsection{An Illustrative Example}
\label{sec:Experiments.IllustrativeExample}

We create a synthetic dataset $\mathcal{D}$ based on the scenario in Fig.~\ref{fig:KarimiV2}. It is a modified version of~\citet[Fig. 1]{Karimi2021_AlgoRecourse}, where we include the protected attribute gender $A$. Here, gender directly affects both an individual's annual salary $X_1$ and bank balance $X_2$, which are used by the bank's ADM $b()$ for approving ($\widehat{Y}=1$) or rejecting ($\widehat{Y}=0$) a loan application. We generate $\mathcal{D}$ for $n=5000$ under $A \sim \text{Ber}(0.45)$ with $A=1$ if the individual is female and $A=0$ otherwise, and assume: $X_1 \leftarrow (-\$1500) \cdot \text{Poi}(10) \cdot A + U_1$; $X_2 \leftarrow (-\$300) \cdot \mathcal{X}^2(4) \cdot A + (3/10) \cdot X_1 + U_2$; and $\widehat{Y} = \mathbbm{1}\{ X_1 + 5 \cdot X_2 > 225000\}$ with $U_1 \sim \$10000 \cdot \text{Poi}(10)$ and $U_2 \sim \$2500 \cdot \mathcal{N}(0, 1)$. Here, $\mathcal{D}$ represents \textit{a known biased scenario}.\footnote{Such ``penalties'', e.g., capture the financial burdens female professionals face in the present after having been discouraged in the past from pursuing high-paying, male-oriented fields \cite{CriadoPerez2019InvisibleWomen}.}  With $A$ we introduce a systematic bias onto the relevant decision attributes for female applicants.

\begin{figure}[t]
\vspace{-2ex}
\begin{minipage}{.45\linewidth}
\begin{figure}[H]
\centering
    \begin{tikzpicture}
        \node (A)  at (-1.75, 0) [circle, draw]{$A$};
        \node (X1) at (0, 0.75) [circle, draw]{$X_1$};
        \node (X2) at (0,-0.75) [circle,draw]{$X_2$};
        \node (Y)  at (1.75, 0) [circle, draw]{$\widehat{Y}$};
        \draw[->] (A) to (X1) {};
        \draw[->] (A) to (X2) {};
        \draw[->] (X1) to (Y) {};
        \draw[->] (X1) to (X2) {};
        \draw[->] (X2) to (Y) {};
    \end{tikzpicture}
\end{figure}
\end{minipage}
\begin{minipage}{.45\linewidth}
\begin{align*}
\mathcal{M} \, & 
\begin{cases}
    A & \leftarrow U_{A} \\
    X_1 & \leftarrow f_1(A)  + U_1 \\
    X_2 & \leftarrow f_2(X_1, A) + U_2
\end{cases}
\end{align*}
\begin{align*}
    \widehat{Y} & = b(X_1, X_2)
\end{align*}
\end{minipage}
\caption{The causal knowledge with corresponding SCM $\mathcal{M}$ and DAG $\mathcal{G}$ behind our (illustrative example) loan application dataset. Let $A$ denote an individual's gender, $X_1$ annual salary, $X_2$ bank balance, and $\widehat{Y}$ the loan decision based on the bank's ADM $b()$.}
\label{fig:KarimiV2}
\end{figure}

\begin{figure}[t]
    \centering
    \begin{subfigure}{.45\linewidth}
    \includegraphics[scale=0.45]{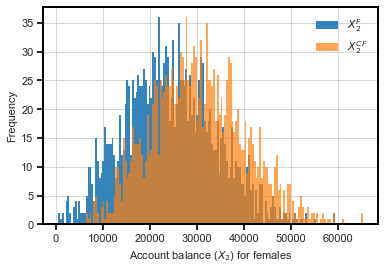}
    \end{subfigure}
    \hfill
    \begin{subfigure}{.45\linewidth}
    \includegraphics[scale=0.45]{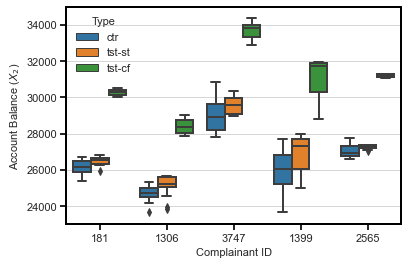}
    \end{subfigure}
\caption{\textbf{Left.} Account balance ($X_2$) distribution for females in the factual $\mathcal{D}$ and counterfactual $\mathcal{D}^{CF}$ datasets. 
\textbf{Right.} A comparison on $X_2$ of the ST and CST (w/o) control group (ctr) versus the ST (tst-st) and CST (w/o) (tst-cf) test groups for five randomly chosen complainants detected by both methods, showing the \textit{fairness given the difference} behind CST as tst-st is closer to ctr than tst-cf.
}
\label{fig:x2_analysis}
\vspace{-3ex}
\end{figure}

To run CST we first generate the counterfactual dataset $\mathcal{D}^{CF}$ based on the intervention $do(A:=0)$, or \textit{what would have happened had all loan applicants been male}? Comparing $\mathcal{D}$ to $\mathcal{D}^{CF}$ already highlights the unwanted systematic effects of $A$. This can be seen, for instance, in Fig.~\ref{fig:x2_analysis} by the rightward shift experienced in $X_2$ for all female applicants when going from the factual to the counterfactual world. 
The loan rejection rate for females drops from 60.9\% in $\mathcal{D}$ to 38.7\% in $\mathcal{D}^{CF}$, which is now closer to the loan rejection rate of 39.2\% experienced by males in both worlds. We run CST for all $k$ sizes. Results are shown in Table~\ref{table:k-results}, where w/o refers to ``without search centers'' for CST.

Does $b()$ discriminate against female applicants? As Table~\ref{table:k-results} shows, all three methods detect a number of individual discrimination cases. On one hand, the bank clearly uses information that is neutral and needed for approving a loan request; on the other hand, this information is tainted by the effects of gender on such information and the bank, in turn, continues to perpetuate biases against women in this scenario. The results show how these neutral provisions are harmful toward female applicants, and uncover potential individual discrimination cases. 

%
\begin{table}[t]
  \caption{Number (and \%) of detected individual discrimination cases for the illustrative example based on gender.}
  \label{table:k-results}
  \centering
  \begin{tabular}{lccccc}
    \toprule
    Method & $k=0$     & $k=15$     & $k=30$ & $k=50$ & $k=100$ \\
    \midrule
    CST (w/o) & 0 & 288 (16.8\%) & 313 (18.3\%) & 342 (20\%) & 395 (23.1\%)  \\
    ST \cite{Thanh_KnnSituationTesting2011} & 0 & 55 (3.2\%) & 65 (3.8\%) & 84 (5\%) & 107 (6.3\%) \\
    CST       & 0 & 420 (24.5\%) & 434 (25.4\%) & 453 (26.5\%) & 480 (28\%)  \\
    CF \cite{Kusner2017CF} & 376 (22\%) &  376 (22\%) &  376 (22\%) &  376 (22\%) & 376 (22\%)  \\
    \bottomrule
  \end{tabular}
\end{table}
%

\textbf{CST relative to situation testing (ST).} Here, consider the CST version without the search centers as ST excludes them. What is clear from Table~\ref{table:k-results} is that CST finds more individual discrimination cases than ST for all $k$ sizes. For $k=50$, e.g., CST (w/o) detects 20\% while ST just 5\%. These results highlight the impact of operationalizing \textit{fairness given the difference}, as the main difference between the two frameworks is how each individual test group is constructed based on the choice of search center. The control group is constructed the same way for both ST and CST.

The choice of the test search centers is what sets CST apart from ST. Note that ST performs an \textit{idealized comparison}. Consider, e.g., the tuple $(x_1=35000, x_2=7948, a=1)$ as the complainant $c$. With $c$ as the test search center, the most similar male profiles to the complainant under any distance $d$ would be tuples similar to $(x_1=35000, x_2=7948, a=0)$. CST, conversely, performs a more \textit{flexible comparison} under \textit{fairness given the difference}. With the corresponding counterfactual tuple $(x_1^{CF}=50796, x_2^{CF}=13852, a^{CF}=0)$ as the test search center, the most similar male profiles to the complainant under the same $d$ would be tuples similar to the counterfactual itself. 
Hence, the test group we construct under CST will probably not be similar to the one we construct under ST nor to the same control group we construct for both ST and CST. Fig.~\ref{fig:x2_analysis}, for instance, shows clearly this result for $k=15$. We randomly chose five complainants that were discriminated by $b()$ according to both ST and CST and plot the distribution of $X_2$ for the control group (ctr), the ST test group (tst-st), and the CST test group (tst-cf). In this scenario, all 55 ST cases are also detected by CST. 


\textbf{CST relative to counterfactual fairness (CF).} Here, consider the CST version including the search centers (though CST w/o is of interest also), as these represent the instances used by CF. We define CF discrimination as a case where the factual $\widehat{y}_c=0$ becomes $\widehat{y}_c^{CF}=1$ after the intervention of $A$. Under this definition, we detect 376 cases of CF discrimination in $\mathcal{D}$, or 22\% of female applicants. CF is independent from $k$ as the framework applies only to the individual comparison of the factual and counterfactual tuples for complainant $c$. Table~\ref{table:k-results} show that CST detects a higher number of individual discrimination cases for each $k$ size (while CST w/o only passes CF at $k=100$). In fact, in this scenario, all cases detected by CF are contained in CST. 

What sets CST apart from CF is twofold. First, CST equips the CF comparison with certainty measures. This point is illustrated in Table~\ref{table:CFwithCIs} where we show individual cases of discrimination detected by both CF and CST along with confidence intervals (CI) \eqref{eq:CIs} provided by the CST framework. Second, CST detects cases of individual discrimination that are counterfactually fair. This point is illustrated in Table~\ref{table:NotinCFwithCIs} where we show individual cases that pass CF but still exhibit a discriminatory pattern when looking at $\Delta p$. Such results highlight why legal stakeholder require multiple comparisons to insure that $c$'s experience is representative of the discrimination claim.

\begin{minipage}[t]{.45\linewidth}
\vspace{-3ex}
\begin{table}[H]
  \caption{Subset of individual discrimination cases detected by both CST ($k=15$) and CF with CI under $\alpha=5\%$}
  \label{table:CFwithCIs}
  \centering
  \begin{tabular}{ccccc}
    \toprule
    Comp. (ID) & $p_c$ & $p_t$ & $\Delta p$ & CI ($\alpha=5$\%)\\
    \midrule
    44  & 1.00 & 0.00 & 1.00* & [1.00, 1.00] \\
    55  & 0.81 & 0.00 & 0.81* & [0.65, 0.97] \\
    150 & 1.00 & 0.94 & 0.06 & [-0.04, 0.16] \\
    203 & 1.00 & 0.88 & 0.13 & [-0.01, 0.26] \\
    218 & 0.56 & 0.00 & 0.56* & [0.36, 0.77] \\
    \bottomrule
  \end{tabular}
\end{table}
\end{minipage}
\hspace{0.55cm}
\begin{minipage}[t]{.45\linewidth}
\vspace{-3ex}
\begin{table}[H]
  \caption{Subset of individual discrimination cases detected by CST ($k=15$) but not by CF with CI under $\alpha=5\%$.}
  \label{table:NotinCFwithCIs}
  \centering
  \begin{tabular}{ccccc}
    \toprule
    Comp. (ID) & $p_c$ & $p_t$ & $\Delta p$ & CI ($\alpha=5$\%)\\
    \midrule
    5    & 0.06 & 0.0 & 0.06 & [-0.04, 0.16] \\
    147  & 0.50 & 0.0 & 0.5* & [0.29, 0.71] \\
    435  & 0.38 & 0.0 & 0.38* & [0.18, 0.58] \\
    1958 & 0.13 & 0.0 & 0.13 & [-0.01, 0.26] \\
    2926 & 0.75 & 0.0 & 0.75* & [0.57, 0.93] \\
    \bottomrule
  \end{tabular}
\end{table}
\end{minipage}
\mbox{}\\[2ex]

\textbf{Confidence in results.} Finally, notice that Tables~\ref{table:CFwithCIs} and \ref{table:NotinCFwithCIs} include cases where $\tau=0$ falls within the individual CI. We detected these cases under $\Delta p > \tau$ (Def.~\ref{def:IndDisc}). Under $\alpha = 5\%$, we would reject these cases as individual discrimination claims with confidence level of 95\% since the minimum deviation is covered by the CIs (Def.~\ref{def:CIs}). These are cases with small $\Delta p$'s that are too close to call. We denote those statistically significant cases with the asterisk on $\Delta p$.
\subsection{Law School Admissions}
\label{sec:Experiments.Real}

Based on the Law School Success example popularized by \citet[Fig. 2]{Kusner2017CF} using US data from the Law School Admission Council survey \cite{Wightman1998_LawDataSource}, we create an admissions scenario to a top law school. We consider as protected attributes an applicant's gender, male/female, (\textit{G}) and race, white/non-white, (\textit{R}). We add an ADM $b()$ that considers the applicant's undergraduate grade-point average (\textit{UGPA}) and law school admissions test scores (LSAT) for admission. If an applicant is successful, $\widehat{Y}=1$; otherwise $\widehat{Y}=0$. We summarize the scenario in Fig.~\ref{fig:LawSchool}. For $b()$, we use the median entry requirements for the top US law school to derive the cutoff $\psi$.\footnote{That being Yale University Law School: \url{https://www.ilrg.com/rankings/law/index/1/asc/Accept}} The cutoff is the weighted sum of 60\% 3.93 over 4.00 in \textit{UGPA} and 40\% 46.1 over 48 \textit{LSAT}, giving a total of 20.8; the maximum possible score under $b()$ is 22 for an applicant. The structural equations follow \eqref{eq:SCM}, as in \cite{Kusner2017CF}, with $b_U$ and $b_L$ denoting the intercepts; $\beta_1$, $\beta_2$, $\lambda_1$, $\lambda_2$ the weights; and $UGPA \sim \mathcal{N}$ and $LSAT \sim \text{Poi.}$ the probability distributions. 

\begin{figure}[h]
\vspace{-2ex}
\begin{minipage}{.35\linewidth}
\begin{figure}[H]
\centering
    \begin{tikzpicture}
        \node (A1)  at (-1.75, -0.55) [circle, draw]{R};
        \node (A2)  at (-1.75, 0.55) [circle, draw]{G};
        \node (X1) at (0, 0.75) [circle, draw]{UGPA};
        \node (X2) at (0,-0.75) [circle,draw]{LSAT};
        \node (Y)  at (1.25, 0) [circle, draw]{$\widehat{Y}$};
        \draw[->] (A1) to (X1) {};
        \draw[->] (A1) to (X2) {};
        \draw[->] (A2) to (X1) {};
        \draw[->] (A2) to (X2) {};
        \draw[->] (X1) to (Y) {};
        \draw[->] (X2) to (Y) {};
    \end{tikzpicture}
\end{figure}
\end{minipage}
\begin{minipage}{.55\linewidth}
\begin{align*}
\mathcal{M} \, & 
\begin{cases}
    R & \leftarrow U_{R}\\
    G & \leftarrow U_G \\
    UGPA & \leftarrow b_U + \beta_1 \cdot R + \lambda_1 \cdot G + U_1, \ \\
    LSAT & \leftarrow  \exp\{b_L + \beta_2 \cdot R + \lambda_2 \cdot G + U_2\}, \\
\end{cases}
\end{align*}
\begin{align*}
    \widehat{Y} & = \mathbbm{1}\{(0.6 \cdot UGPA + 0.4 \cdot LSAT) > \psi\}
\end{align*}
\end{minipage}
\caption{The causal knowledge with corresponding SCM $\mathcal{M}$ and DAG $\mathcal{G}$ behind the law school admissions dataset, with $R$ denoting race ($R=1$ for non-white) and $G$ denoting gender ($G=1$ for female).}
\label{fig:LawSchool}
\vspace{-2ex}
\end{figure}

The dataset $\mathcal{D}$ contains $n=21790$ applicants, 43.8\% are females and 16.1\% are non-whites. Despite the ADM $b()$ being externally imposed by us for the purpose of illustrating the CST framework, under $b()$ only 1.88\% of the female applicants are successful compared to 2.65\% of the male applicants; similarly, only 0.94\% of the non-white applicants are successful compared to 2.58\% of the white applicants. Therefore, is $b()$ discriminatory toward non-white and female applicants? We run CST along with ST and CF for each protected attribute. We generate the counterfactual dataset $\mathcal{D}^{CF}$ for each $R$--\textit{what would have been the outcome had all applicants been white}?--and $G$--\textit{what would have been the outcome had all applicants been male}?--and present the results, respectively, in Table~\ref{table:k-results_RACE} and Table~\ref{table:k-results_GENDER}. Both tables show CST detecting more individual cases of discrimination than ST and CF. 
%

%
\begin{table}[H]
\vspace{-1ex}
  \caption{Number (and \%) of individual discrimination cases for the law school admissions scenario based on race.}
  \label{table:k-results_RACE}
  \centering
  \begin{tabular}{lccccc}
    \toprule
    Method & $k=0$     & $k=15$     & $k=30$ & $k=50$ & $k=100$ \\
    \midrule
    CST (w/o) & 
        0 & 256 (7.3\%) & 309 (8.81\%) & 337 (9.61\%) & 400 (11.41\%)  \\
    ST \cite{Thanh_KnnSituationTesting2011} & 
        0 & 33 (0.94\%) & 51 (1.45\%) & 61 (1.74\%) & 64 (1.83\%) \\
    CST & 
        0 & 286 (8.16\%) & 309 (8.81\%) & 337 (9.61\%) & 400 (11.41\%)  \\
    CF \cite{Kusner2017CF} &
        231 (6.59\%) &  231 (6.59\%) &  231 (6.59\%) &  231 (6.59\%) & 231 (6.59\%)  \\
    \bottomrule
  \end{tabular}
\end{table}
\begin{table}[H]
\vspace{-3ex}
\caption{Number (and \%) of individual discrimination cases in for the law school admissions scenario based on gender.}
  \label{table:k-results_GENDER}
  \centering
  \begin{tabular}{lccccc}
    \toprule
    Method & $k=0$     & $k=15$     & $k=30$ & $k=50$ & $k=100$ \\
    \midrule
    CST (w/o) & 
        0 & 78 (0.82\%) & 120 (1.26\%) & 253 (2.65\%) & 296 (3.10\%)  \\
    ST \cite{Thanh_KnnSituationTesting2011} & 
        0 & 77 (0.81\%) & 101 (1.06\%) & 229 (2.4\%) & 258 (2.71\%) \\
    CST & 
        0 & 99 (1.04\%) & 129 (1.35\%) & 267 (2.80\%) & 296 (3.10\%)  \\
    CF \cite{Kusner2017CF} &
        56 (0.59\%) &  56 (0.59\%) &  56 (0.59\%) &  56 (0.59\%) & 56 (0.59\%)  \\
    \bottomrule
  \end{tabular}
\end{table}

A similar analysis of Tables~\ref{table:k-results_RACE} and \ref{table:k-results_GENDER} for CST (w/o) versus ST and CST versus CF as in Section~\ref{sec:Experiments.IllustrativeExample} follows. What the results here highlight, though, is how the two versions of CST compare to each other. In both tables, as $k$ increases, CST (w/o) catches on to CST in the number of cases. This is likely related to how the observations for female/male and non-white/white are distributed in the dataset; though, it also relates to the fact that the difference in size between the groups in each version is just one instance: $k$ versus $k+1$. We should observe the same trend in Table~\ref{table:k-results} if we continued to increase $k$. These results show that the different runs of CST can eventually reach the same conclusions under a certain $k$ size. In practice, it means that we could implement one of the two versions of CST without compromising the number of detected individual discrimination cases, though further research is needed.  

%
%

\section{Conclusion}
\label{sec:Conclusion}
We presented counterfactual situation testing (CST), a new framework for detecting individual discrimination in a dataset of classifier decisions. 
Compared to other methods, CST uncovers more cases even when the classifier is counterfactually fair. It also equips counterfactual fairness with uncertainty measures. 
CST acknowledges the pervasive effects of the protected attribute by comparing individual instances in the dataset that are observably different in the factual world but hypothetically similar in the counterfactual world.
Thus, the results are not too surprising as CST operationalizes \textit{fairness given the difference}, which is a more flexible take on similarity between individuals for testing discrimination than the standard, idealized comparison of two individuals that only differ on their protected status.

{\bf Implementation.} CST is, above all, a framework for detecting discrimination that advocates for building the test group on the generated counterfactual of the complainant. How similarity is defined, e.g., obviously conditions the implementation. We presented a k-NN version with $d$ as \eqref{eq:Distance}; other implementations are possible and still loyal to CST as long as the construction of the control and test groups follows \textit{fairness given the difference}. 

Similarly, detecting discrimination is a difficult, context-specific task. That is why for CST we emphasized the role of the expert in constructing the causal graph necessary for generating the counterfactual instances. Indeed, this step could be optimized using, e.g., causal discovery methods \cite{Peters2017_CausalInference}, but proving discrimination is time consuming and should remain as such given its sensitive role in our society. Further, we are aware that, e.g., the experimental setting could be pushed further by considering higher dimensions or more complex causal structures. What is the point in doing so, though, if that is not the case with current ADM tools being deployed and audited in real life like the recent Dutch scandal \cite{Heikkila2022_DutchScnadal}? Proving discrimination is not a problem exclusive to (causal) modeling. With CST we wanted to create a framework aware of the multiple angles to the problem of proving discrimination. The cases we have tackled here are intended to showcase what is possible implementation-wise. 

{\bf Limitations.} As future work, promising directions include extending the framework to cases where causal sufficiency does not hold, which is a common risk, and to cases where the decision maker $b()$ is non-binary or of a specific type (e.g., a decision tree). Here, we have also focused on tabular data. Future work should push CST further into more complex datasets to explore the scalability and robustness of the framework.

{\bf Research Ethics.} In this work we used anonymized benchmark (and synthetic) data designed for public use, complying with applicable legal and ethical rules. We disclosed all details of our method in line with the transparency mandates of the ACM Code of Ethics. 

%
%

\begin{acks}
    Many thanks to Alejandra Bringas and Ioanna Papageorgiou from NoBIAS for their legal commentary. This work has received funding from the European Union’s Horizon 2020 research and innovation program under Marie Sklodowska-Curie Actions (grant agreement number 860630) for the project "NoBIAS - Artificial Intelligence without Bias". This work reflects only the authors' views and the European Research Executive Agency (REA) is not responsible for any use that may be made of the information it contains. 
\end{acks}

\bibliographystyle{ACM-Reference-Format}
\bibliography{references}

\newpage
\appendix
\label{sec:Appendix}
\section{Supplementary Material}

\subsection{Working example for generating counterfactuals}
\label{A:ExampleCounterfactuals}

We present a simple working example for counterfactual generation. Given the assumptions we undertake for \eqref{eq:SCM} plus the additional assumption of an additive noise model (ANM)---i.e., $\mathcal{S} = \{X_j \leftarrow f_j(X_{pa(j)}) + U_j \}_{j=1}^p$---the generating procedure is straightforward. The ANM assumption is also assumed in Section~\ref{sec:Experiments} for the classification scenarios. It is a common model specification assumption that allows to identify more easily the non-random parts of the equation. Suppose we have the following structural causal model $\mathcal{M}$ and corresponding directed acyclical graph $\mathcal{G}$:

\begin{minipage}{.45\linewidth}
\begin{figure}[H]
\centering
    \begin{tikzpicture}
        \node (X1)  at (-1.65, 0) [circle, draw]{$X_1$};
        \node (X2) at (0, 0.85) [circle, draw]{$X_2$};
        \node (X3) at (1.65,0) [circle,draw]{$X_3$};
        \draw[->] (X1) to (X2) {};
        \draw[->] (X1) to (X3) {};
        \draw[->] (X2) to (X3) {};
    \end{tikzpicture}
\end{figure}
\end{minipage}
\begin{minipage}{.45\linewidth}
\begin{align*}
\mathcal{M} \, & 
\begin{cases}
    X_1 & \leftarrow U_1 \\
    X_2 & \leftarrow \alpha \cdot X_1  + U_2 \\
    X_3 & \leftarrow \beta_1 \cdot X_1 + \beta_2 \cdot X_2 + U_3
\end{cases}
\end{align*}
\end{minipage}
\medskip

\noindent
where $U_1, U_2, U_3$ represent the latent variables, $X_1, X_2, X_3$ the observed variables, and $\alpha, \beta_1, \beta_2$ the coefficient for the causal effect of, respectively, $X_1 \rightarrow X_2$,  $X_1 \rightarrow X_3$, and $X_2 \rightarrow X_3$. Suppose we want to generate the counterfactual for $X_3$, i.e., $X_3^{CF}$, had $X_1$ been equal to $x_1 \in X_1$. In the \textit{abduction} step, we estimate $U_1$, $U_2$, and $U_3$ given the evidence or what is observed under the specified structural equations:
\begin{align*}
    \hat{U}_1 & = X_1 \\
    \hat{U}_2 & = X_2 - \alpha \cdot X_1 \\
    \hat{U}_3 & = X_3 - \beta_1 \cdot X_1 + \beta_2 \cdot X_2
\end{align*}
We can generalize this step for \eqref{eq:SCM} as $U_j = X_j - f_j(X_{pa(j)})$ $\forall X_j \in X$. This step is an individual-level statement on the residual variation under SCM $\mathcal{M}$. It accounts for all that our assignment functions $f_j$, which are at the population level, cannot explain, or the \textit{error terms}. 
In the \textit{action} step, we intervene $X_1$ and set all of its instances equal to $x_1$ via $do(X_1:=x_1)$ and obtaining the intervened DAG $\mathcal{G}'$ and SCM $\mathcal{M}'$:

\begin{minipage}{.45\linewidth}
\begin{figure}[H]
\centering
    \begin{tikzpicture}
        \node (X1)  at (-1.65, 0) [circle, draw]{$do(x_1)$};
        \node (X2) at (0, 0.85) [circle, draw]{$X_2$};
        \node (X3) at (1.65,0) [circle,draw]{$X_3$};
        \draw[->] (X2) to (X3) {};
    \end{tikzpicture}
\end{figure}
\end{minipage}
\begin{minipage}{.45\linewidth}
\begin{align*}
\mathcal{M}' \, & 
\begin{cases}
    X_1 & = x_1 \\
    X_2 & \leftarrow \alpha \cdot x_1  + U_2 \\
    X_3 & \leftarrow \beta_1 \cdot x_1 + \beta_2 \cdot X_2 + U_3
\end{cases}
\end{align*}
\end{minipage}
\medskip

\noindent
where no edges come out from $X_1$ as it has been fixed to $x_1$. Finally, in the \textit{prediction} step, we combine these two steps to calculate $X_3^{CF}$ under the set of $\hat{U}$ and the intervened $\mathcal{M}'$:
\begin{align*}
    X_3^{CF} & \leftarrow \beta_1 \cdot x_1 + \beta_2 \cdot X_2 + \hat{U}_3 \\
             & \leftarrow \beta_1 \cdot x_1 + \beta_2 \cdot (\alpha \cdot x_1 + \hat{U}_2) + \hat{U}_3
\end{align*}
which is done for all instances in $X_3$. This is what is done at a larger scale, for example, in \cite{Karimi2021_AlgoRecourse} and \cite{Pearl2016_CausalInference}, and also in this paper. The same three steps can apply to $X_2$ (also for $X_1$, though it would be trivial as it is a root note).

We can view this approach as a \textit{frequentist}\footnote{This is not a formal distinction, but based on talks with other researchers in counterfactual generation. Such a distinction, to the best of our knowledge, remains an open question.} one for generating counterfactuals, in particular, with regard to the Abduction step. A more \textit{Bayesian} approach is what is done by \cite{Kusner2017CF} where they use a Monte Carlo Markov Chain (MCMC) to draw $\hat{U}$ by updating its prior distribution with the evidence $X$ to then proceed with the other two steps. In Section~\ref{sec:Experiments.Real}, we used both approaches for generating the counterfactuals and found no difference in the results. We only present in this paper the first approach as it is less computationally expensive. 

\subsection{Sketch of Proof for Proposition~\ref{prop:ActioanbleCF}}
\label{A:ProofActionableCF}

Consider the factual tuple $(x_c, a_c=1, \widehat{y}_c=0)$ and assume the generated counterfactual is $(x_c^{CF}, a_c^{CF}=0, \widehat{y}_c^{CF}=0)$. Since $\widehat{y}_c = \widehat{y}_c^{CF}$, this is a case where counterfactual fairness holds. However, the decision boundary of the model $b()$ can be purposely set such that the $k$-nearest neighbors of $x_c$ are all within the decision $\hat{Y}=0$, and less than $1-\tau$ fraction of the $k$-nearest neighbors of $x_c^{CF}$ are within the decision $\hat{Y}=0$. This leads to a $\Delta p > 1-(1-\tau) = \tau$, showing that there is individual discrimination. The other way can be shown similarly by assuming $\widehat{y}_c \neq \widehat{y}_c^{CF}$ but the sets of $k$-nearest neighbors have rates of negative decisions whose difference is lower than $\tau$.

\section{Algorithms for k-NN CST Implementation}
\label{A:Algorithms}

We present the relevant algorithms for the k-NN CST implementation (Section~\ref{sec:CST.AnImplementation}). The algorithm~\ref{alg:run_cfST} performs CST while algorithm~\ref{alg:get_topk} returns the indices of the top-$k$ tuples with respect to the search centers based on the distance function $d$. Notice that the main difference in algorithm~\ref{alg:run_cfST} when creating the neighborhoods is that the search centers are drawn from the factual dataset for the control group $\mathcal{D}$ and the counterfactual dataset $\mathcal{D}^{CF}$ for the test group. Further, notice that we use the same $c$ (i.e., index) for both as these two data-frames have the same structure by construction. 

\newcommand\mycommfont[1]{\scriptsize\ttfamily\scriptsize{#1}}
\SetCommentSty{mycommfont}

\IncMargin{1.5em}
\begin{algorithm2e}[h!]
\small 
    \caption{run\_CST}
    \label{alg:run_cfST}
	\SetKwInOut{Input}{Input}
	\SetKwInOut{Output}{Output}
	\Input{$\mathcal{D}$, $\mathcal{D}^{CF}$, $k$}
	\Output{$[p_c - p_t]$}
	\BlankLine
	    $prot\_condition \leftarrow \mathcal{D}[:, \; prot\_attribute] == prot\_value$\\
	    $\mathcal{D}_c \leftarrow \mathcal{D}[prot\_condition]$\tcp*[f]{get protected (control) search space}\\
	    $\mathcal{D}_t \leftarrow \mathcal{D}[\neg \; prot\_condition]$\tcp*[f]{get non-protected (test) search space}\\
	    $prot\_idx \leftarrow \mathcal{D}_c.index.to\_list(\;)$\tcp*{get idx for all complainants}
	    $diff\_list = [ \; ]$ \\
	    \For{$c, \; row \in prot\_idx$}{
        $res\_1 \leftarrow get\_top\_k(\mathcal{D}\; \; \; \; \; [c, \; :], \mathcal{D}_c, k)$\tcp*{idx of the top-k tuples for control group} 
        $res\_2 \leftarrow get\_top\_k(\mathcal{D}^{CF}[c, \; :], \mathcal{D}_t, k)$\tcp*{idx of the top-k tuples for test group} 
        $p_c \leftarrow sum(\mathcal{D}[res_1, \; target\_attribute]==negative\_outcome) \; / \; len(res\_1)$ \\
        $p_t \leftarrow sum(\mathcal{D}[res_2, \; target\_attribute]==negative\_outcome) \; / \; len(res\_2)$ \\
        $diff\_list[c] \leftarrow p_c - p_t$}
    \Return{$diff\_list$}
\end{algorithm2e}

\IncMargin{1.5em}
\begin{algorithm2e}[h!]
\small 
    \caption{get\_top\_k}
    \label{alg:get_topk}
	\SetKwInOut{Input}{Input}
	\SetKwInOut{Output}{Output}
	\Input{$t$, $t\_set$, $k$}
	\Output{$[indices]$}
	\BlankLine
     $(idx, dist) \leftarrow k\_NN(t, t\_set, k + 1)$\tcp*{run k-NN algorithm with $k+1$}
    \If{without search centers} {
        $remove(t, idx, dist)$\tcp*{remove the center t from idx}
    }
    $idx' \leftarrow sort(idx, dist)$\tcp*{sort idx by the distance}
    \Return{$idx'$}
\end{algorithm2e}

\end{document}